\pdfoutput=1
\documentclass[11pt]{article}

\usepackage[final]{acl}
\usepackage{acl}
\usepackage{times}
\usepackage{latexsym}
\usepackage[T1]{fontenc}
\usepackage[utf8]{inputenc}
\usepackage{microtype}
\usepackage{inconsolata}

\usepackage{graphicx}
\usepackage{multirow}
\usepackage{booktabs} % 引入booktabs包来使用加粗线命令

\usepackage{tabularx} 
\usepackage{float}

\usepackage{amsmath} 
\usepackage{makecell}
\usepackage{array} % 支持 p{} 环境
\usepackage{longtable}

\usepackage{amssymb}

\usepackage[linesnumbered,ruled,vlined]{algorithm2e}

\title{Ontology-Guided Reverse Thinking Makes Large Language Models Stronger on Knowledge Graph Question Answering}

\author{
  \textbf{Runxuan Liu\textsuperscript{1}},
  \textbf{Bei Luo\textsuperscript{2}},
  \textbf{Jiaqi Li\textsuperscript{3,4}\thanks{Corresponding author.}},
  \textbf{Baoxin Wang\textsuperscript{1,3}},
  \textbf{Ming Liu\textsuperscript{1,5}\thanks{Corresponding author.}},\\
  \textbf{Dayong Wu\textsuperscript{3}},
  \textbf{Shijin Wang\textsuperscript{3}},
  \textbf{Bing Qin\textsuperscript{1,5}}\\
  \textsuperscript{1}Harbin Institute of Technology, Harbin, China\\
  \textsuperscript{2}Beijing University of Posts and Telecommunications, Beijing, China\\
  \textsuperscript{3}Joint Laboratory of HIT and iFLYTEK, Beijing, China \\
  \textsuperscript{4}University of Science and Technology of China, Hefei, China\\
  \textsuperscript{5}Pengcheng Laboratory, Shenzhen, China\\
  \quad\texttt{\{rxliu,mliu,qinb\}@ir.hit.edu.cn},
  \quad\texttt{luobei@bupt.edu.cn}
}

\begin{document}
\maketitle

\begin{abstract}
Large language models (LLMs) have shown remarkable capabilities in natural language processing. However, in knowledge graph question answering tasks (KGQA), existing methods rely on entity vector matching, but the purpose of the question is abstract and difficult to match with specific entities. As a result, it is difficult to efficiently establish reasoning paths to the purpose, which leads to information loss and redundancy. To address this issue, inspired by human reverse thinking, we propose Ontology-Guided Reverse Thinking (ORT), a novel framework that constructs reasoning paths from purposes back to conditions. ORT operates in three key phases: (1) using LLM to extract purpose labels and condition labels, (2) constructing label reasoning paths based on the KG ontology, and (3) using the label reasoning paths to guide knowledge retrieval. Experiments on the WebQSP and CWQ datasets show that ORT achieves state-of-the-art performance and significantly enhances the capability of LLMs for KGQA. Our code is publicly available at \url{https://github.com/Runxuan-Liu/ORT}.
\end{abstract}

\section{Introduction}
LLMs have made significant achievements in natural language processing, excelling in tasks such as semantic understanding \cite{Raiaan2024ARO}, text generation \cite{shen-etal-2024-heart}, machine translation \cite{hu-etal-2024-gentranslate}, dialogue systems \cite{Zhang2019DIALOGPTL}, sentiment analysis \cite{li2025perception}, and text summarization \cite{Basyal2023TextSU}. LLMs have also been applied in various scenarios, such as the medical field \cite{Wu2024MedicalGR} and scientific research support \cite{Wu2024SparkRAAR}.

\begin{figure}[t]
  \centering
  \includegraphics[width=\columnwidth]{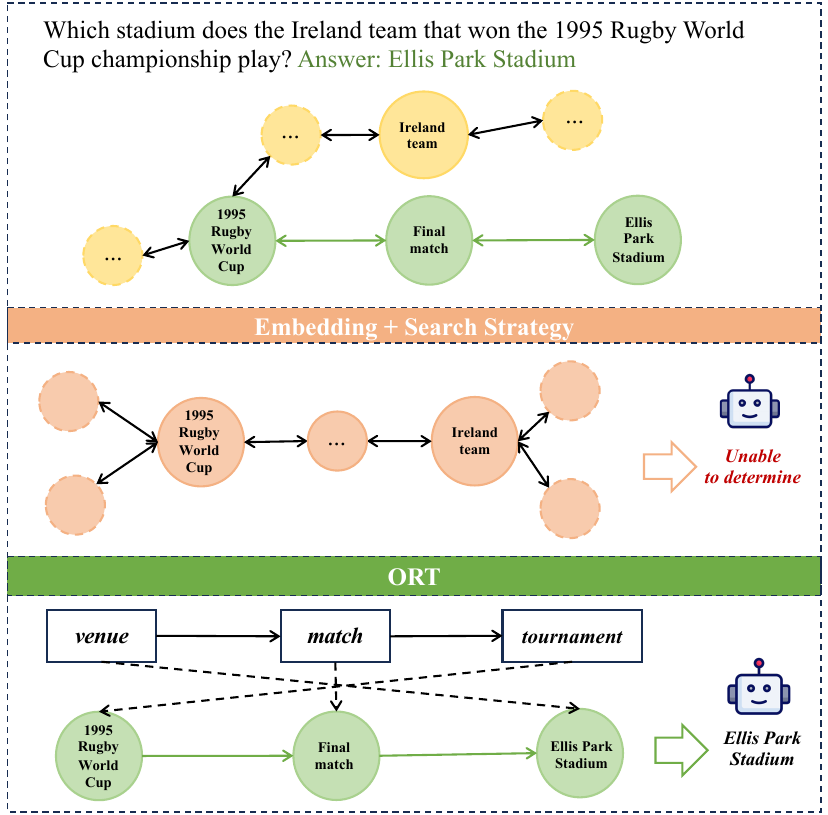}  % 设置图片宽度，保证其占用一栏
  \caption{Example of previous methods and ORT. Traditional methods are limited to entity-centric reasoning through vector matching and path collection. In contrast, ORT enables ontology-aware reasoning by identifying conceptual intents, constructing reverse-label reasoning paths, and guiding targeted traversal to the correct answers.}  % 图片标题
  \label{fig:introduce}  % 图片引用标签
\end{figure}

\begin{figure*}[t]  % 'h'表示图片插入位置
  \centering
  \includegraphics[width=1\textwidth]{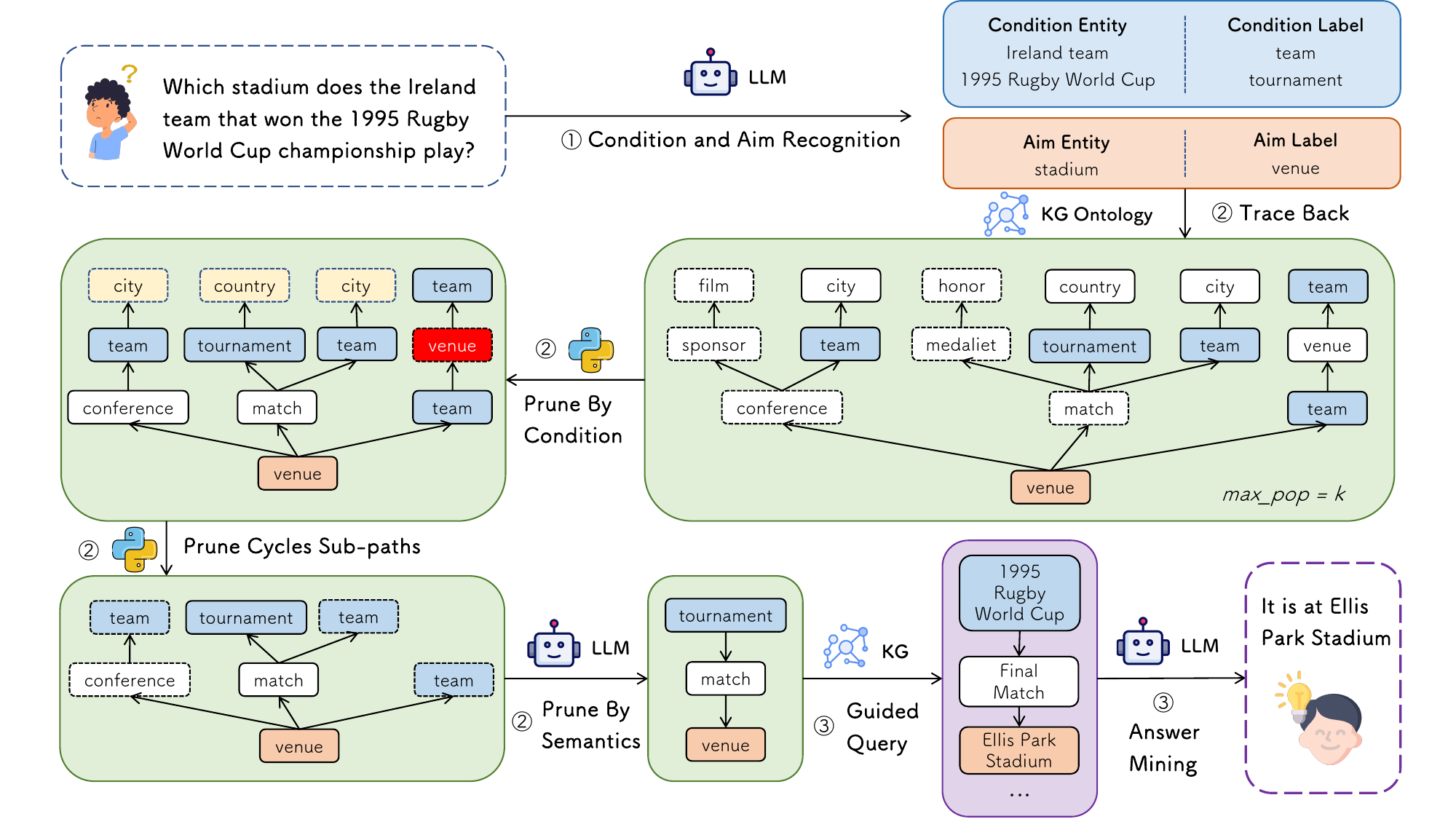}  % 设置图片宽度，并替换为图片文件名
  \caption{The overall framework of ORT. Starting from a question, the LLM is used to identify conditions and aims in the question, along with their corresponding labels. Using the aim label as the root node, the system iteratively queries related labels on the knowledge graph ontology until the backward max-hop limit is reached. Paths that do not contain condition labels, paths after the last condition label in each sequence, and loops are then pruned. The reasoning paths are used as guidance to query the knowledge graph, and the LLM summarizes the entity paths to derive the final answer.}  % 图片标题
  \label{fig:framework}  % 图片引用标签
\end{figure*}

The rapid development of LLMs has sparked interest in combining LLMs with knowledge graphs to improve KGQA performance \cite{Hu2024GRAGGR}. 
Existing approaches typically adopt two paradigms. The first is \textit{fine-tuning methods}, such as LPKG~\cite{wang-etal-2024-learning-plan} and RoG~\cite{Luo2023ReasoningOG}. 
However, creating high-quality training data is resource-intensive \cite{Cao2023InstructionMI}. Additionally, knowledge graphs are highly structured data, and when faced with questions that have not been fine-tuned, the quality is difficult to guarantee \cite{Jiang2024KGFITKG}. 
The second is \textit{embedding + search methods}, such as MindMap~\cite{wen-etal-2024-mindmap} and Think-on-Graph~\cite{Sun2023ThinkonGraphDA}, which rely on entity embeddings and graph traversal but are unable to handle conceptual targets absent in KG entities. As shown in Figure ~\ref{fig:introduce}, ``stadium'' is a concept, not an entity in the knowledge graph, so ``Embedding + Search Strategy'' can only find paths between ``1995 Rugby World Cup'' and ``Ireland Team'' and their neighbors, but cannot reach ``Ellis Park Stadium''.

In this paper, we propose a novel method named Ontology-guided Reverse Thinking (ORT). Our approach begins by extracting not only known entities from the question but also its underlying aims using LLMs, where these aims are represented as entity labels. Building upon these labels, we establish a reasoning framework that combines both ontological structures and reverse thinking principles. Specifically, we construct a reasoning tree that originates from the identified aims and progresses toward the known conditions, effectively creating label reasoning paths with the knowledge graph ontology. This reverse-oriented approach incorporates path pruning, eliminating unnecessary branches during the reasoning process. The refined reasoning paths then guide targeted knowledge queries in the knowledge graph, followed by using LLMs to aggregate knowledge and generate answers. This integrated methodology enables precise knowledge retrieval while minimizing interference from irrelevant information, ultimately enhancing the accuracy of the LLM's responses.

The experimental results demonstrate that our method significantly improves the answer coverage and quality of LLM KGQA. Compared to direct responses from LLMs, our method achieves a Hit@1 improvement of at least 25.43\% and an F1 score improvement of at least 25.82\%. As a plug-and-play approach, it greatly enhances the efficiency of LLM KGQA.

In summary, our main contributions include: 1) We first introduce a human-like problem-solving approach for KGQA: Ontology-Guided Reverse Thinking. 2) As a plug-and-play method, we enable LLMs to efficiently understand the structure of the knowledge graph. 3) Experimental results demonstrate a significant improvement in the LLM's KGQA ability, achieving state-of-the-art among the models studied.

\section{Methodology}
As shown in Figure~\ref{fig:framework}, the entire algorithm is divided into three steps:

\begin{enumerate}
    \item \textbf{Condition and Aim Recognition}: Prompt the LLM to understand the known conditions and the solving aims of the question.
    \item \textbf{Ontology-Guided Reverse Thinking Reasoning}:  Use the Reverse Thinking Reasoning method to construct label reasoning paths on the knowledge graph ontology.
    \item \textbf{Guided Answer Mining}: Use the label reasoning paths to guide querying and prompt the LLM to generate the final answer.
\end{enumerate}

\subsection{Aim and Condition Recognition}

\begin{figure}[htbp]
  \centering
  \includegraphics[width=\columnwidth]{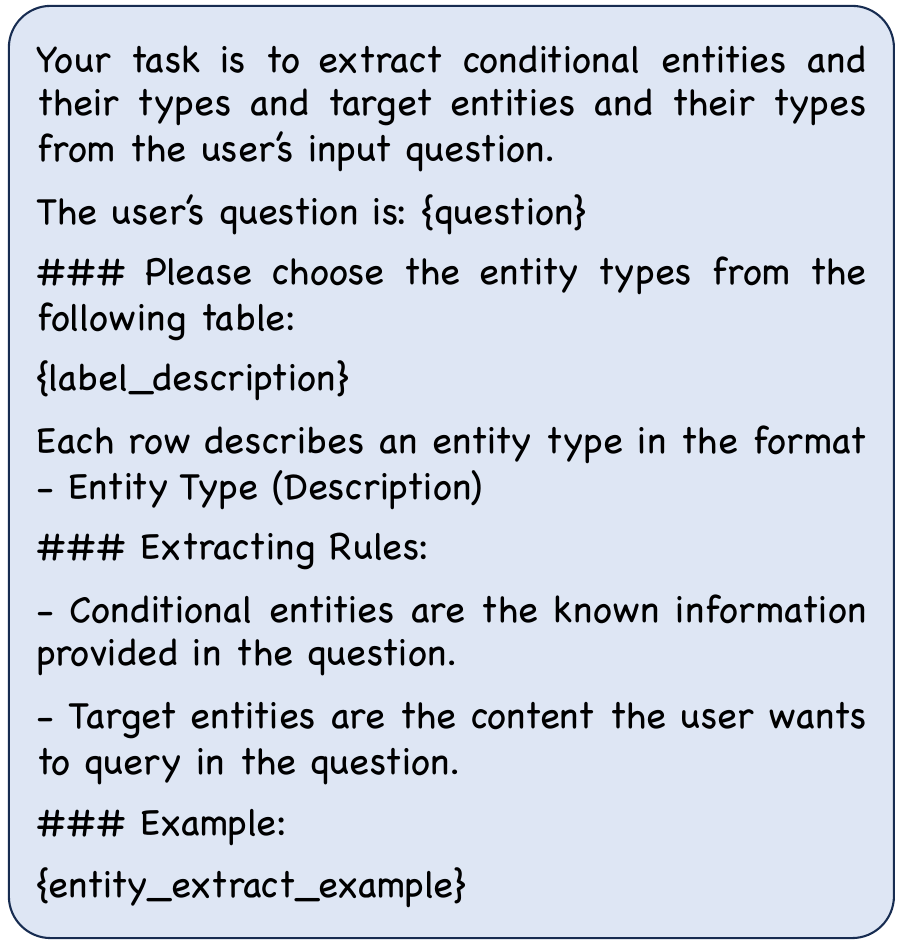}  % 设置图片宽度，保证其占用一栏
  \caption{Prompt template for condition and aim recognition.}  % 图片标题
  \label{fig:extract_prompt}  % 图片引用标签
\end{figure}

This step extracts the condition entities \( \mathcal{C}_E = \{c_1, c_2, \dots, c_n\} \), labels of condition entities \( \mathcal{C}_L = \{cl_1, cl_2, \dots, cl_n\} \), aim entities \( \mathcal{A}_E = \{a_1, a_2, \dots, a_n\} \), and labels of aim entities \( \mathcal{A}_L = \{al_1, al_2, \dots, al_n\} \) from the question by prompting LLM. 

Condition is defined as the known key information in the question, while aim is defined as the content the user wants to query through the question. The aim entity refers to the entity in the user question that conveys the intended purpose. In fact, the aim entity is not used in the subsequent processing steps. It serves as an intermediate step for obtaining its corresponding labels. The aim entity labels represent a set of related labels of the aim entity in the knowledge graph, play a crucial role in establishing a mapping from the user question to the knowledge graph ontology. This label identification process effectively addresses the limitations of relying solely on vector-based matching.

We provide the LLM with a Label List of the knowledge graph, prompting the LLM to first extract \( \mathcal{C}_E \) and \( \mathcal{A}_E \), and then assign labels to the respective entities. The main content of the prompt template is shown in Figure~\ref{fig:extract_prompt}, and the complete content of the prompt can be found in Appendix~\ref{sec:PROMPTS}.

\subsection{Ontology-Guided Reverse Thinking Reasoning}

Knowledge graph reasoning differs from document reasoning in that its data is structured, making the effective use of structural information particularly important \cite{Thambi2022TowardsIT}. We propose for the first time the use of a knowledge graph ontology (KG ontology) to construct label reasoning paths, thereby guiding KG queries to enhance the reasoning ability of LLM with knowledge graphs. 

The way we construct paths on KG ontology is Reverse Thinking Reasoning. The constructed path is named as label reasoning path \( \mathcal{R}_P \), or abstract reasoning path.

\subsubsection{Step I. Construct the Neighbor Label Dictionary}

The ontology of the knowledge graph consists of several relation-defined triples. For each label \( l_i \) in a triple, we collect all other labels \( l_k, l_{k+1}, \dots \) that appear in the same relation-defined triple.

To express this, we introduce a function \( \mathcal{N}(l_i) \), which denotes the set of labels \( l_k \) that appear in the same triple as \( l_i \):

\begin{equation}
\mathcal{N}(l_i) = \{ l_k \mid (l_i, \text{relationship}, l_k) \in \mathcal{G} \}
\end{equation}
where \( \mathcal{G} \) represents the set of all triples in the knowledge graph. Then, we construct a neighbor label dictionary, denoted as \( \mathcal{D} \), where \( l_i \) is the key, and \( \mathcal{N}(l_i) \) is the value associated with it:

\begin{equation}
\mathcal{D} = \{ l_i : \mathcal{N}(l_i) \}
\end{equation}

For example, given the following relationships in the knowledge graph ontology: \( l_1 \) → \( l_2 \), \( l_1 \) → \( l_3 \), and \( l_3 \) → \( l_1 \), the neighbor label dictionary \( \mathcal{D} \) will be:

\[
\mathcal{D} =
\left\{
\begin{array}{ll}
l_1 & : [l_2, l_3], \\
l_2 & : [l_1], \\
l_3 & : [l_1]
\end{array}
\right.
\]

\subsubsection{Step II. Construct the Reverse Reasoning Tree}

First, since the length of \( \mathcal{A}_L \) may be greater than 1, we create a virtual root node and add all aim labels \( \mathcal{A}_L = \{ al_1, al_2, \dots, al_m \} \) as its child nodes.

Then, we recursively traverse all child nodes, querying the neighbor label dictionary \( \mathcal{D} \) to add all neighboring labels \( \mathcal{N}(l_i) \) as child nodes of each current node. 

This process continues recursively until the maximum recursion depth \( \text{max\_pop} \) is reached. The maximum recursion depth is determined based on the number of hops of the question. 

The reverse reasoning tree, denoted as \( \mathcal{T} \), is built as a recursive structure where the nodes represent labels from the knowledge graph, and the edges represent the relationships between them. Due to the limited space in the image, the relationships between the entity labels are not explicitly shown.

\subsubsection{Step III. Prune By Conditions}

Starting from the root node, we perform a depth-first search (DFS) and record the current path.

When a leaf node is reached, we check if any node in the path matches the condition labels \( \mathcal{C}_L \). The pruning is performed as follows:

\begin{itemize}
    \item If the path contains no condition label nodes, the entire path is removed.
    \item If the path contains condition label nodes, only the last condition node and its preceding nodes are retained, while the subsequent nodes are deleted.
\end{itemize}

The pruning algorithm is recursively applied to each child node, using a copy of the path to avoid contaminating the original path. The output is the tree after pruning by conditions, denoted as \( \mathcal{T}_{\mathcal{\text{Condition}}} \).

\begin{algorithm}[htbp]
\caption{Prune Paths by Conditions}
\KwIn{$\mathcal{R}_P \leftarrow \text{all label reasoning paths by DFS}$, $\mathcal{C}_L \leftarrow \text{condition labels}$}
\KwOut{$\mathcal{T}_{\mathcal{\text{Condition}}}$}

\SetKwFunction{FMain}{PrunePathsByConditions}
\SetKwProg{Fn}{Function}{:}{}
\Fn{\FMain{$\mathcal{R}_P$, $\mathcal{C}_L$}}{
    $\mathcal{T}_{\mathcal{\text{Condition}}} \gets \emptyset$\;
    
    \ForEach{$path \in \mathcal{R}_P$}{
        $condition\_indices \gets \left[ i \mid node_i \in path \text{ and } node_i \in \mathcal{C}_L \right]$\;
        
        \If{$condition\_indices \neq \emptyset$}{
            $last\_condition\_index \gets \text{last element of } condition\_indices$\;
            $\mathcal{T}_{\mathcal{\text{Condition}}}.\text{append}(path[:last\_condition\_index + 1])$\;
        }
    }
    \KwRet{$\mathcal{T}_{\mathcal{\text{Condition}}}$}\;
}
\end{algorithm}

\subsubsection{Step IV. Prune Cycle Sub-paths}

Due to the possibility of bidirectional relationships between two labels, cycles may exist in the reasoning paths. 

A depth-first search (DFS) is performed on \( \mathcal{T}_{\text{Condition}} \), adding the current node's name to the visited set \( \text{visited} \). The pruning algorithm is recursively called for each child node of the current node. If the current node's name already exists in the visited set \( \text{visited} \), the edge between the current node and its parent is removed, effectively eliminating the cycle. During backtracking, the current node's name is removed from the visited set so that other paths can access it. The output is the tree after pruning cycles, denoted as \( \mathcal{T}_{\text{Cycle}} \).

\subsubsection{Step V. Prune By Semantics}

As shown in Figure~\ref{fig:framework}, after pruning by conditions and cycles, interference paths such as ``\( \text{team} \rightarrow \text{conference} \rightarrow \text{venue} \)'' may still exist. To remove these irrelevant paths, semantic information is used for pruning. 

A depth-first search (DFS) is performed on all paths of \( \mathcal{T}_{\text{Cycle}} \), and the paths are reversed to forward paths. These paths, together with the problem, are input into LLM. The model is prompted using a template to output the paths that are beneficial for answering the question. The main content of the prompt template is shown in Figure~\ref{fig:filter_prompt}, and the complete content can be found in Appendix~\ref{sec:PROMPTS}.

\begin{figure}[H]
  \centering
  \includegraphics[width=\columnwidth]{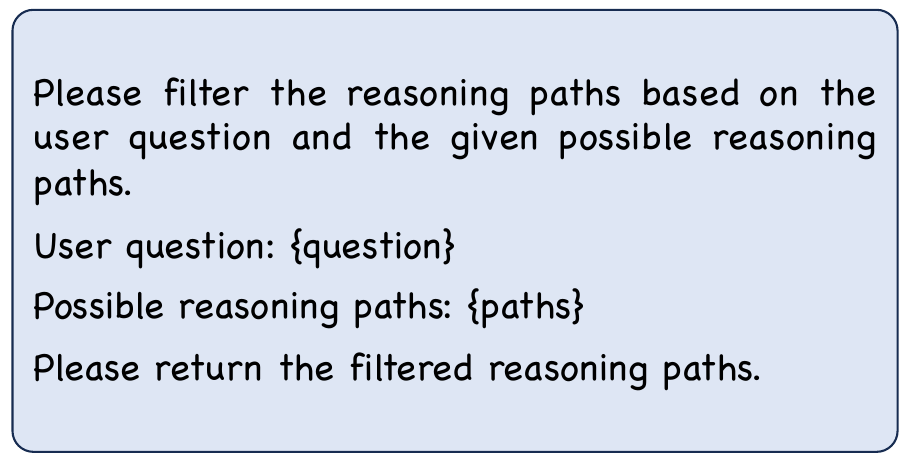}  % 设置图片宽度，保证其占用一栏
  \caption{Prompt template for prune by semantics with LLM.}  % 图片标题
  \label{fig:filter_prompt}  % 图片引用标签
\end{figure}

\subsection{Guided Answer Mining}

Through Ontology-Guided Reverse Thinking Reasoning, abstract reasoning paths for solving the problem are obtained. They are then used to guide the forward knowledge graph query process to collect entity reasoning paths. 

A tree structure is used to store the results of each query step. The process is driven by traversing the abstract path, which consists of a sequence of labels. For each reasoning path, the first node is a condition node, and all entities that satisfy the label of this node are added as child nodes to the tree. Then, for each of these child nodes, the next label in the abstract path is used to query the neighboring entities of the current entity. Only those neighbors whose label matches the next label in the abstract path are retained and added as children of the current child node. This process continues iteratively, following the order of labels in the abstract path.

\begin{table*}[t]
  \centering
  \normalsize
  \caption{The result of our method and other baseline methods on the WebQSP dataset and the CWQ dataset.}
  \label{table:baseline}
  \renewcommand{\arraystretch}{1.2} % 调整行高
  \setlength{\tabcolsep}{6pt} % 调整列间距
  \begin{tabular}{l p{5cm} c c c c}
    \toprule
    \multirow{2}{*}{Type} & \multirow{2}{*}{Method} & \multicolumn{2}{c}{WebQSP} & \multicolumn{2}{c}{CWQ} \\
    \cmidrule(lr){3-4} \cmidrule(lr){5-6}
                          &                                          & Hit@1  & F1     & Hit@1 & F1 \\
    \midrule
    \multirow{2}{*}{Embedding} 
                          & KV-Mem \cite{Miller2016KeyValueMN}           & 46.7   & 34.5   & 18.4  & 15.7 \\
                          & NSM \cite{He2021ImprovingMK}                 & 68.7   & 62.8   & 47.6  & 42.4 \\
    \midrule
    \multirow{3}{*}{Retrieval} 
                          & GraftNet \cite{Sun2018OpenDQ}                & 66.4   & 60.4   & 36.8  & 32.7 \\
                          & SR+NSM \cite{Zhang2022SubgraphRE}            & 68.9   & 64.1   & 50.2  & 47.1 \\
                          & SR+NSM+E2E \cite{Zhang2022SubgraphRE}        & 69.5   & 64.1   & 49.3  & 46.3 \\
    \midrule
    \multirow{3}{*}{LLMs} 
                          & GPT-4o                  & 61.8   & 43.6   & 38.2  & 32.9 \\
                          & Qwen-max                & 59.0   & 40.0   & 36.4  & 29.5 \\
                          & DeepSeek-v3 \cite{DeepSeekAI2024DeepSeekV3TR}             & 64.0   & 43.9   & 41.1  & 33.8 \\
    \midrule
    \multirow{2}{*}{LLM+KGs(Fine-tuned)} 
                          & KD-CoT \cite{Wang2023KnowledgeDrivenCE}                 
                          & 68.6   & 52.5   & 55.7  & -    \\
                          & RoG \cite{Luo2023ReasoningOG}                    
                          & 85.7   & 70.8   & 62.6  & 56.2 \\
    \midrule
    \multirow{3}{*}{LLM+KGs(non-Fine-tuned)} 
                          & KG Retriever            
                          & 63.0   & 42.9   & 46.7  & 40.2 \\
                          & MindMap \cite{wen-etal-2024-mindmap}        
                          & 64.9   & 47.1   & 48.8  & 43.3\\
                          \cmidrule(lr){2-6} % 添加额外的水平线
                          & ORT             & \textbf{89.4} & \textbf{71.8} & \textbf{72.9} & \textbf{62.6} \\
    \bottomrule
  \end{tabular}
\end{table*}

If there are many neighboring entities satisfying the next label, and the number exceeds the limit, top\_k neighboring entities are randomly selected and added to the tree. This process is recursively applied until the reasoning path is fully traversed.

After the forward entity tree is built, a depth-first search (DFS) is performed to collect all entity paths, which are then input into LLM along with the problem to generate the final answer. The complete content of the prompt can be found in Appendix~\ref{sec:PROMPTS}.

\section{Experiments}
\subsection{Experimental Setup}

\paragraph{Benchmarks}

We conducted experiments on two widely used KGQA datasets: WebQuestionSP (WebQSP) \cite{Yih2016TheVO} and ComplexWebQuestions (CWQ) \cite{Talmor2018TheWA}. Both datasets are constructed by extracting data from the Freebase knowledge graph. In our experiments, we follow RoG \cite{Luo2023ReasoningOG} to construct knowledge graphs for WebQSP and CWQ. More details can be found in Appendix~\ref{sec:EXPERIMENT DETAILS}. 

\begin{figure}[t]
  \centering
  \includegraphics[width=\columnwidth]{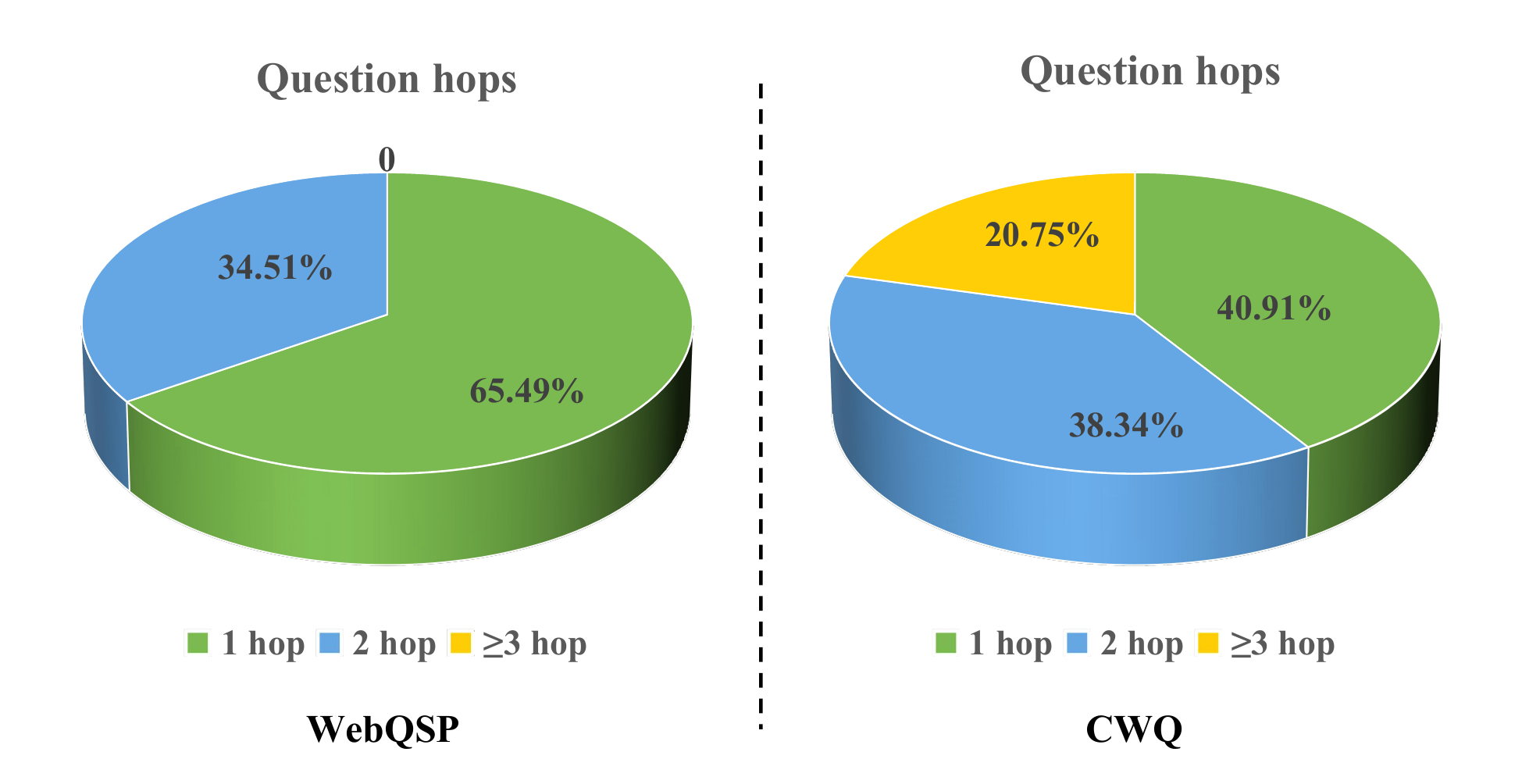}  % 设置图片宽度为两栏宽
  \caption{Statistics of the question hops in WebQSP and CWQ.}  % 图片标题
  \label{fig:question hops}  % 图片引用标签
\end{figure}

\begin{figure*}[t]
  \centering
  \includegraphics[width=\textwidth]{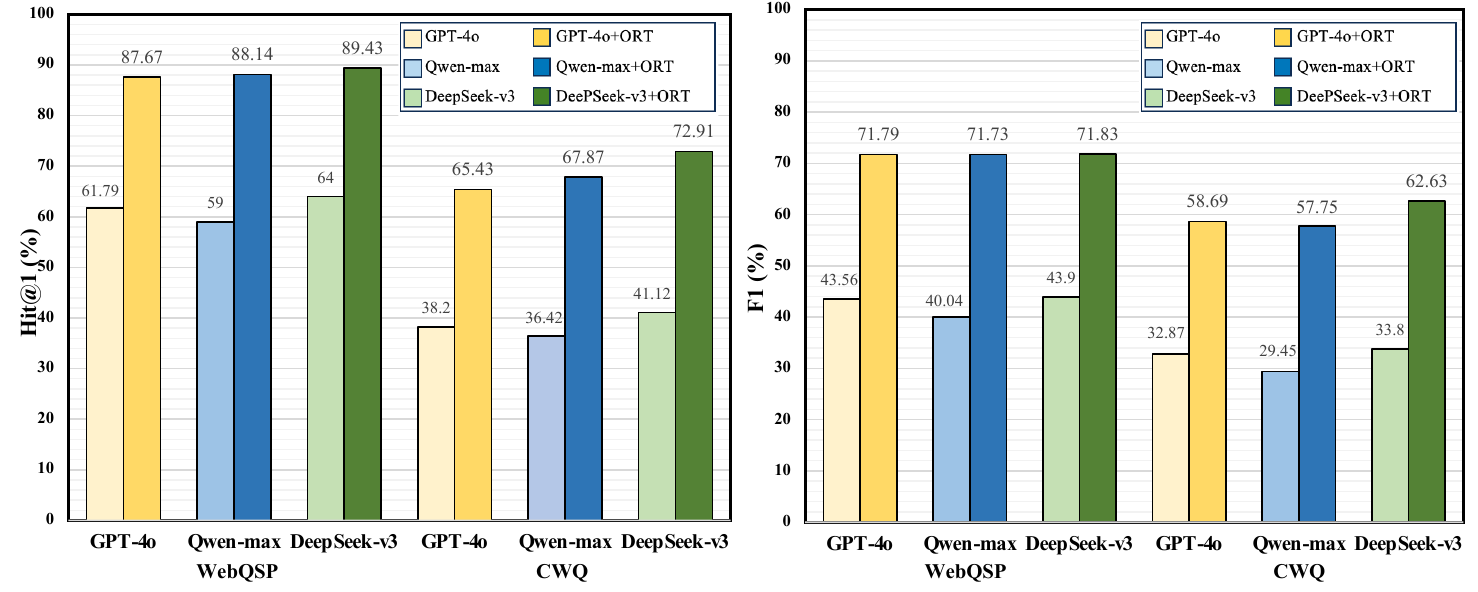}  % 设置图片宽度为两栏宽
  \caption{Comparison of LLM vs. LLM+Our Method on WebQSP and CWQ Datasets. The left side is Hit@1 (\%), and the right side is F1 (\%).}  % 图片标题
  \label{fig:LLM Improvement}  % 图片引用标签
\end{figure*}

\paragraph{Evaluation Metrics}

Following previous work \cite{Luo2023ReasoningOG, Zhang2022SubgraphRE, Li2024SimpleIE, Tan2024PathsoverGraphKG}, we use Hit@1 and F1 as evaluation metrics. We also provide detailed results for accuracy, precision, and recall in Appendix~\ref{sec:ADDTIONAL RESULTS}. Hit@1 refers to selecting the top-1 prediction and checking whether the true answer is included. If yes, the score is 1; otherwise, it is 0. This measures the answer coverage. Since the prediction may contain invalid content, the F1 score is also used as an evaluation metric to balance precision and recall.

\paragraph{Baselines}

We compared a total of five categories of baselines, including:
\begin{itemize}
    \item \textbf{Embedding-based}: KV-Mem \cite{Miller2016KeyValueMN} and NSM \cite{He2021ImprovingMK};
    \item \textbf{Retrieval-based}: GraphNet \cite{Sun2018OpenDQ} and SR \cite{Zhang2022SubgraphRE};
    \item \textbf{LLM}: GPT-4, DeepSeek-v3 \cite{DeepSeekAI2024DeepSeekV3TR}, and Qwen-max;
    \item \textbf{Fine-tuned LLM knowledge graph reasoning methods (KGR FT)}: KD-CoT \cite{Wang2023KnowledgeDrivenCE} and RoG \cite{Luo2023ReasoningOG};
    \item \textbf{Non-fine-tuned LLM knowledge graph reasoning methods (KGR w/o FT)}: MindMap \cite{wen-etal-2024-mindmap} and KG-Retriever.
    % and Think-on-Graph (ToG) \cite{Sun2023ThinkonGraphDA}.
\end{itemize}
Details of these baselines can be found in Appendix~\ref{sec:EXPERIMENT DETAILS}.

\subsection{ORT Achieves SOTA}

As shown in Table~\ref{table:baseline}, ORT achieves state-of-the-art performance on both WebQSP and CWQ. The base LLM for all LLM+KGs(non-Fine-tuned) methods including our method shown in Table~\ref{table:baseline} is DeepSeek-v3.

Compared to small-scale models based on embedding or retrieval, on WebQSP, Hit@1 improves by 20\% to 42.7\%, and F1 improves by 7.7\% to 37.3\%; on CWQ, Hit@1 improves by 22.7\% to 54.5\%, and F1 improves by 15.5\% to 46.9\%. 

ORT also outperforms partially \textit{KGR FT} methods such as KD-CoT and RoG. This demonstrates that our method not only enhances question-answering performance but also reduces costs, improves model scalability, and enhances adaptability to different knowledge graphs. 

Besides, compared to pure LLM and \textit{KGR w/o FT} methods, ORT also achieves significant improvements, which will be discussed later.

Furthermore, as shown in Figure ~\ref{fig:question hops}, WebQSP primarily focuses on single-hop questions, with 65.49\% requiring only one hop and no questions exceeding three hops, whereas CWQ contains more complex multi-hop questions, with 20.75\% requiring three or more hops. The inferior performance of current methods on CWQ compared to WebQSP further underscores the limitations of existing approaches in addressing multi-hop reasoning tasks.

\subsection{Soars LLM's KGQA Ability}

As shown in Figure~\ref{fig:LLM Improvement}, we conducted comparative experiments on three LLMs: GPT-4o, DeepSeek-v3, and Qwen-Max. Compared to direct answers, LLMs using ORT led to more than a 25\% improvement in Hit@1 and F1 scores across both datasets. 

On WebQSP, Hit@1 of the three LLM respectively improved by 25.7\%, 25.3\%, and 29.14\%, and F1 score increased by 28.23\%, 27.96\%, and 31.69\%. On CWQ, Hit@1 improved by 27.23\%, 31.79\%, and 31.45\%, and F1 score increased by 25.82\%, 28.83\%, and 28.30\%. Details can be found in Table~\ref{tabel:llm-improvement}.

This not only demonstrates that ORT effectively enhances the performance of LLMs on KGQA tasks but also highlights that ORT is not limited to a specific LLM. It serves as a universal enhancement strategy and can be directly used for improving LLM performance. It has the potential to become an effective, convenient, and important tool in such a domain.

\begin{table*}[htbp]
  \centering
  \normalsize
  \caption{Comparison of LLM vs. LLM+Our Method on WebQSP and CWQ Datasets}
  \label{tabel:llm-improvement}
  \renewcommand{\arraystretch}{1.2} % 减少行高
  \setlength{\tabcolsep}{5pt} % 减少列间距
  \begin{tabularx}{\textwidth}{l *{4}{>{\centering\arraybackslash}X}}
    \toprule
    \multirow{2}{*}{Method} & \multicolumn{2}{c}{WebQSP} & \multicolumn{2}{c}{CWQ} \\
    \cmidrule(lr){2-3} \cmidrule(lr){4-5}
                            & Hit@1  & F1     & Hit@1 & F1 \\
    \midrule
    GPT-4o                   & 61.79    & 43.56    & 38.20   & 32.87 \\
    GPT-4o + ORT      & \textbf{87.67} \textcolor[HTML]{009900}{(↑25.7)} & \textbf{71.79} \textcolor[HTML]{009900}{(↑28.23)} & \textbf{65.43} \textcolor[HTML]{009900}{(↑27.23)} & \textbf{58.69} \textcolor[HTML]{009900}{(↑25.82)} \\
    \midrule
    QWen-max                 & 59.00    & 40.04    & 36.42   & 29.45 \\
    QWen-max + ORT    & \textbf{88.14} \textcolor[HTML]{009900}{(↑29.14)} & \textbf{71.73} \textcolor[HTML]{009900}{(↑31.69)} & \textbf{67.87} \textcolor[HTML]{009900}{(↑31.45)} & \textbf{57.75} \textcolor[HTML]{009900}{(↑28.3)} \\
    \midrule
    DeepSeek-v3              & 64.0    & 43.9    & 41.12   & 33.80 \\
    DeepSeek-v3 + ORT & \textbf{89.43} \textcolor[HTML]{009900}{(↑25.43)} & \textbf{71.83} \textcolor[HTML]{009900}{(↑27.93)} & \textbf{72.91} \textcolor[HTML]{009900}{(↑31.79)} & \textbf{62.63} \textcolor[HTML]{009900}{(↑28.83)} \\
    \bottomrule
  \end{tabularx}
\end{table*}

\subsection{ORT Outperforms Peers}

% ORT effectively addresses two key shortcomings of previous \textit{KGR w/o FT} methods: "Embedding + Search Strategy" \cite{wen-etal-2024-mindmap} and over-reliance on LLM to generate reasoning paths \cite{Chen2024ANP}. 

\begin{table}[h]
\centering
\caption{The results of non-Fine-Tuned LLM KG Reasoning methods on WebQSP and CWQ}
\label{tabel:non-fine-tuned-llm-kgr}
\footnotesize % 调整字体大小
\renewcommand{\arraystretch}{1.2} % 调整表格行高
\setlength{\tabcolsep}{4pt} % 调整列间距
\begin{tabularx}{\linewidth}{l*{4}{>{\centering\arraybackslash}X}} % 所有列居中对齐
\toprule
\multirow{2}{*}{Method} & \multicolumn{2}{c}{WebQSP} & \multicolumn{2}{c}{CWQ} \\
\cmidrule(lr){2-3} \cmidrule(lr){4-5}
 & Hit@1 & F1 & Hit@1 & F1 \\
\midrule
ORT + GPT-4o      & \textbf{87.67} & \textbf{71.79} & \textbf{65.43} & \textbf{58.69} \\
MindMap + GPT-4o         & 61.17 & 46.09 & 51.33 & 44.84 \\
KG Retriever + GPT-4o    & 60.15 & 42.44 & 46.67 & 41.14 \\
\hline
ORT + DeepSeek-v3 & \textbf{89.43} & \textbf{71.83} & \textbf{72.91} & \textbf{62.63} \\
MindMap + DeepSeek-v3    & 64.92 & 47.14 & 48.83 & 43.30 \\
KG Retriever + DeepSeek-v3 & 63.01 & 42.87 & 47.67 & 40.20 \\
\hline
ORT + QWen-max    & \textbf{88.14} & \textbf{71.73} & \textbf{67.87} & \textbf{57.75} \\
MindMap + QWen-max       & 59.46 & 43.31 & 45.50 & 40.35 \\
KG Retriever + QWen-max  & 57.16 & 39.91 & 45.00 & 38.99 \\
\bottomrule
\end{tabularx}
\end{table}

ORT achieves better performance compared to other methods of the same type. Using the same base LLM (GPT-4, DeepSeek-v3, and Qwen-max), ORT was compared with MindMap and KG Retriever, as shown in Table~\ref{tabel:non-fine-tuned-llm-kgr}. On WebQSP, ORT achieved an average improvement of 26.56\% in Hit@1 and 26.27\% in F1 over MindMap across the three base models. On CWQ, ORT outperformed MindMap with an average improvement of 20.18\% in Hit@1 and 16.86\% in F1. 

\subsection{Ablation Study}
\begin{table*}[t]
\centering
\caption{Ablation Experiment Results}
\label{tabel:ablation}
\begin{tabular}{c c c c c | c c c c }
\toprule
\multirow{2}{*}{Method} & \multicolumn{4}{c|}{WebQSP} & \multicolumn{4}{c}{CWQ} \\
\cmidrule(lr){2-5} \cmidrule(lr){6-9}
 & Hit@1 & Precision & Recall & F1 & Hit@1 & Precision & Recall & F1 \\
\midrule
ORT & \textbf{89.43} & \textbf{80.92} & 74.51 & 71.83 & \textbf{72.91} & \textbf{65.57} & \textbf{66.03} & \textbf{62.63} \\
w/o LLM Filter & 86.58 & 75.10 & \textbf{78.54} & \textbf{73.01} & 62.58 & 57.45 & 54.76 & 53.24 \\
Trace Forward & 77.82 & 71.92 & 61.91 & 58.73 & 60.73 & 51.50 & 54.18 & 49.30 \\
w/o Rules & 64.00 & 56.91 & 46.55 & 43.87 & 41.12 & 36.42 & 36.40 & 33.80 \\
\bottomrule
\end{tabular}
\end{table*}
Through ablation study, we aim to demonstrate the effectiveness and necessity of Reverse Thinking Reasoning, Knowledge Graph Structure-Based Reasoning, and Rule-Guided Reasoning. We designed it in three parts:
\begin{enumerate}
    \item \textbf{w/o LLM Filter}: In this part, we removed using LLMs for pruning based on the semantics of questions and paths.
    \item \textbf{Trace Forward}: We additionally designed a forward reasoning algorithm, which collects reasoning paths starting from the conditions and iterating towards the goals.
    \item \textbf{w/o Rules}: In this part, label reasoning paths are not constructed, and instead, LLM directly generates answers.
\end{enumerate}

The experimental results are shown in Table~\ref{tabel:ablation}, including Hit@1, F1, Precision, and Recall. Precision measures the proportion of correct predictions among all predicted results, while Recall measures the proportion of correct predictions identified from all ground truth instances.

\paragraph{w/o LLM Filter Analysis}As seen in Table~\ref{tabel:ablation}, on WebQSP, Hit@1 and Precision decreased, but Recall and F1 improved. To ensure the reliability of predictions and reduce hallucinations, we still choose to retain the use of LLM to filter abstract paths.

\paragraph{Trace Forward Analysis}To contrast with Trace Back, we designed the Trace Forward method, which performs forward reasoning starting from the conditions. The basic pipeline is to extract the conditions from the question, and then iteratively perform a breadth-first search from several conditions on the KG ontology to construct a reasoning tree. A depth-first search from the conditions outputs all paths, which, along with the question, are given to LLM to filter abstract paths that semantically match. Then, abstract paths are used to retrieve entity paths from the knowledge graph, which are passed to LLM to generate the final answer. 

The potential drawback of this method is that it may collect irrelevant abstract paths, and overly depends on LLM to generate reasoning paths. As seen in Table~\ref{tabel:ablation}, on WebQSP, Trace Forward's Hit@1 decreased by 11.61\%, and F1 decreased by 13.10\%. One CWQ, Trace Forward's Hit@1 decreased by 12.18\%, and F1 decreased by 13.33\%. However, this method still performed better than MindMap, which highlights the necessity of utilizing the KG ontology and using abstract paths to guide knowledge retrieval.

\paragraph{w/o Rules Analysis}Finally, to demonstrate the necessity of Rule-Guided Reasoning, we conducted experiments without rule guidance, i.e., directly using LLM to generate answers without generating abstract paths. The experimental results show that this method's performance significantly decreased.

\section{Related Work}
\paragraph{Small-scale models for KGQA}

Small-scale methods for knowledge graph question answering (KGQA) can be divided into two categories: embedding-based and retrieval-based methods. Embedding-based methods, such as KV-Mem \cite{Miller2016KeyValueMN} and NSM \cite{He2021ImprovingMK}, represent entities and relations in a low-dimensional vector space, performing well on simple, single-hop queries. However, they struggle with complex, multi-hop queries due to difficulty in capturing intricate path information. To address this, retrieval-based models like GraphNet \cite{Sun2018OpenDQ} and SR \cite{Zhang2022SubgraphRE} construct subgraphs or paths for reasoning, showing improvements in multi-hop tasks by better leveraging structural relationships. Yet, both methods are limited by incomplete utilization of the full structural information in the knowledge graph.

\paragraph{Fine-tuning LLMs for KGQA}

In recent years, the rapid development of large language models (LLMs) has sparked interest in combining LLMs with knowledge graphs to improve KGQA performance. Models like RoG \cite{Luo2023ReasoningOG}, KD-CoT \cite{Wang2023KnowledgeDrivenCE}, UniKGQA \cite{Jiang2022UniKGQAUR}, and DeCAF \cite{Yu2022DecAFJD} have demonstrated impressive results by fine-tuning LLMs to generate reasoning paths and produce answers. These models excel at tackling complex KGQA tasks, where multi-hop reasoning is required. However, fine-tuning LLMs often demands vast computational resources and labeled datasets, making it challenging to scale these methods for practical, real-world applications.

\paragraph{Non-fine-tuning LLMs for KGQA}

Some recent approaches focus on methods that utilize LLMs for KGQA without requiring additional training. MindMap \citet{wen-etal-2024-mindmap} is one such method that extracts entities from the query and performs a breadth-first search in the knowledge graph to generate reasoning paths. Additionally, \citet{Chen2024ANP} proposes a model that feeds all the relations in the knowledge graph to the LLM to help generate relational paths. Although these methods can avoid the high computational costs associated with fine-tuning, they often suffer from a lack of deep understanding of the underlying structure of the knowledge graph, which can lead to the generation of lower-quality reasoning paths.

\section{Conclusion}
In this paper, we simulate the cognitive paradigm that humans use to solve complex problems and propose the Ontology-Guided Reverse thinking method for knowledge graph question answering. We use LLMs to understand the intent of the question and generate the corresponding labels. By leveraging the knowledge graph ontology, we use Reverse-Thinking Reasoning to form label reasoning paths, followed by guided knowledge graph queries and answer aggregation. Experimental results show that our method significantly improves the accuracy and answer coverage.

\section*{Limitations}
For this work, we want to address two areas for improvement in the future. First, when querying the knowledge graph along reasoning paths, there may be a large number of entities satisfying the label constraints, which could lead to irrelevant results. Second, when generating the final answer, inputting all entity paths into the LLMs may introduce irrelevant paths, potentially lowering the accuracy of the answer.

\section*{Acknowledgement}
The research in this article is supported by the New Generation of Artificial Intelligence National Science and Technology Major Project (2023ZD0121503),  the National Science Foundation of China (U22B2059, 62276083).

\bibliography{references}

@inproceedings{Yih2016TheVO,
  title={The Value of Semantic Parse Labeling for Knowledge Base Question Answering},
  author={Wen-tau Yih and Matthew Richardson and Christopher Meek and Ming-Wei Chang and Jina Suh},
  booktitle={Annual Meeting of the Association for Computational Linguistics},
  year={2016},
  url={https://api.semanticscholar.org/CorpusID:13905064}
}

@article{Talmor2018TheWA,
  title={The Web as a Knowledge-Base for Answering Complex Questions},
  author={Alon Talmor and Jonathan Berant},
  journal={ArXiv},
  year={2018},
  volume={abs/1803.06643},
  url={https://api.semanticscholar.org/CorpusID:3986974}
}

@article{Luo2023ReasoningOG,
  title={Reasoning on Graphs: Faithful and Interpretable Large Language Model Reasoning},
  author={Linhao Luo and Yuan-Fang Li and Gholamreza Haffari and Shirui Pan},
  journal={International Conference on Learning Representations},
  year={2024},
  volume={abs/2310.01061},
  url={https://api.semanticscholar.org/CorpusID:263605944}
}

@inproceedings{Zhang2022SubgraphRE,
  title={Subgraph Retrieval Enhanced Model for Multi-hop Knowledge Base Question Answering},
  author={Jing Zhang and Xiaokang Zhang and Jifan Yu and Jian Tang and Jie Tang and Cuiping Li and Hong Chen},
  booktitle={Annual Meeting of the Association for Computational Linguistics},
  year={2022},
  url={https://api.semanticscholar.org/CorpusID:247158305}
}

@article{Li2024SimpleIE,
  title={Simple is Effective: The Roles of Graphs and Large Language Models in Knowledge-Graph-Based Retrieval-Augmented Generation},
  author={Mufei Li and Siqi Miao and Pan Li},
  journal={ArXiv},
  year={2024},
  volume={abs/2410.20724},
  url={https://api.semanticscholar.org/CorpusID:273654355}
}

@article{Tan2024PathsoverGraphKG,
  title={Paths-over-Graph: Knowledge Graph Empowered Large Language Model Reasoning},
  author={Xingyu Tan and Xiaoyang Wang and Qing Liu and Xiwei Xu and Xin Yuan and Wenjie Zhang},
  journal={ArXiv},
  year={2024},
  volume={abs/2410.14211},
  url={https://api.semanticscholar.org/CorpusID:273482196}
}

@article{Miller2016KeyValueMN,
  title={Key-Value Memory Networks for Directly Reading Documents},
  author={Alexander H. Miller and Adam Fisch and Jesse Dodge and Amir-Hossein Karimi and Antoine Bordes and Jason Weston},
  journal={ArXiv},
  year={2016},
  volume={abs/1606.03126},
  url={https://api.semanticscholar.org/CorpusID:2711679}
}

@article{He2021ImprovingMK,
  title={Improving Multi-hop Knowledge Base Question Answering by Learning Intermediate Supervision Signals},
  author={Gaole He and Yunshi Lan and Jing Jiang and Wayne Xin Zhao and Ji-rong Wen},
  journal={Proceedings of the 14th ACM International Conference on Web Search and Data Mining},
  year={2021},
  url={https://api.semanticscholar.org/CorpusID:231572861}
}

@article{Sun2018OpenDQ,
  title={Open Domain Question Answering Using Early Fusion of Knowledge Bases and Text},
  author={Haitian Sun and Bhuwan Dhingra and Manzil Zaheer and Kathryn Mazaitis and Ruslan Salakhutdinov and William W. Cohen},
  journal={ArXiv},
  year={2018},
  volume={abs/1809.00782},
  url={https://api.semanticscholar.org/CorpusID:52154304}
}

@techreport{DeepSeekAI2024DeepSeekV3TR,
  title={DeepSeek-V3 Technical Report},
  author={{DeepSeek-AI} and Aixin Liu and Bei Feng and Bing Xue and Bing-Li Wang},
  institution={DeepSeek-AI},
  year={2024},
  url={https://api.semanticscholar.org/CorpusID:275118643}
}

@article{Wang2023KnowledgeDrivenCE,
  title={Knowledge-Driven CoT: Exploring Faithful Reasoning in LLMs for Knowledge-intensive Question Answering},
  author={Keheng Wang and Feiyu Duan and Sirui Wang and Peiguang Li and Yunsen Xian and Chuantao Yin and Wenge Rong and Zhang Xiong},
  journal={ArXiv},
  year={2023},
  volume={abs/2308.13259},
  url={https://api.semanticscholar.org/CorpusID:261214582}
}

@inproceedings{wen-etal-2024-mindmap,
    title = "{M}ind{M}ap: Knowledge Graph Prompting Sparks Graph of Thoughts in Large Language Models",
    author = "Wen, Yilin  and
      Wang, Zifeng  and
      Sun, Jimeng",
    editor = "Ku, Lun-Wei  and
      Martins, Andre  and
      Srikumar, Vivek",
    booktitle = "Proceedings of the 62nd Annual Meeting of the Association for Computational Linguistics (Volume 1: Long Papers)",
    month = aug,
    year = "2024",
    address = "Bangkok, Thailand",
    publisher = "Association for Computational Linguistics",
    url = "https://aclanthology.org/2024.acl-long.558/",
    doi = "10.18653/v1/2024.acl-long.558",
    pages = "10370--10388",
}

@inproceedings{Chen2024ANP,
  title={A New Pipeline for Knowledge Graph Reasoning Enhanced by Large Language Models Without Fine-Tuning},
  author={Zhongwu Chen and Long Bai and Zixuan Li and Zhen Huang and Xiaolong Jin and Yong Dou},
  booktitle={Conference on Empirical Methods in Natural Language Processing},
  year={2024},
  url={https://api.semanticscholar.org/CorpusID:273901282}
}

@article{Thambi2022TowardsIT,
  title={Towards Improving the Performance of Question Answering System using Knowledge Graph - A Survey},
  author={Sincy V. Thambi and P. C. Reghuraj},
  journal={2022 Second International Conference on Artificial Intelligence and Smart Energy (ICAIS)},
  year={2022},
  pages={672-679},
  url={https://api.semanticscholar.org/CorpusID:247795830}
}

@article{Jiang2022UniKGQAUR,
  title={UniKGQA: Unified Retrieval and Reasoning for Solving Multi-hop Question Answering Over Knowledge Graph},
  author={Jinhao Jiang and Kun Zhou and Wayne Xin Zhao and Ji-rong Wen},
  journal={ArXiv},
  year={2022},
  volume={abs/2212.00959},
  url={https://api.semanticscholar.org/CorpusID:254221022}
}

@article{Yu2022DecAFJD,
  title={DecAF: Joint Decoding of Answers and Logical Forms for Question Answering over Knowledge Bases},
  author={Donghan Yu and Shenmin Zhang and Patrick Ng and Henghui Zhu and Alexander Hanbo Li and J. Wang and Yiqun Hu and William Wang and Zhiguo Wang and Bing Xiang},
  journal={ArXiv},
  year={2022},
  volume={abs/2210.00063},
  url={https://api.semanticscholar.org/CorpusID:252683172}
}

@inproceedings{Devlin2019BERTPO,
  title={BERT: Pre-training of Deep Bidirectional Transformers for Language Understanding},
  author={Jacob Devlin and Ming-Wei Chang and Kenton Lee and Kristina Toutanova},
  booktitle={North American Chapter of the Association for Computational Linguistics},
  year={2019},
  url={https://api.semanticscholar.org/CorpusID:52967399}
}

@inproceedings{Bollacker2008FreebaseAC,
  title={Freebase: a collaboratively created graph database for structuring human knowledge},
  author={Kurt D. Bollacker and Colin Evans and Praveen K. Paritosh and Tim Sturge and Jamie Taylor},
  booktitle={SIGMOD Conference},
  year={2008},
  url={https://api.semanticscholar.org/CorpusID:207167677}
}

@article{Raiaan2024ARO,
  title={A Review on Large Language Models: Architectures, Applications, Taxonomies, Open Issues and Challenges},
  author={Mohaimenul Azam Khan Raiaan and Md. Saddam Hossain Mukta and Kaniz Fatema and Nur Mohammad Fahad and Sadman Sakib and Most. Marufatul Jannat Mim and Jubaer Ahmad and Mohammed Eunus Ali and Sami Azam},
  journal={IEEE Access},
  year={2024},
  volume={12},
  pages={26839-26874},
  url={https://api.semanticscholar.org/CorpusID:267675587}
}

@inproceedings{shen-etal-2024-heart,
    title = "{HEART}-felt Narratives: Tracing Empathy and Narrative Style in Personal Stories with {LLM}s",
    author = "Shen, Jocelyn  and
      Mire, Joel  and
      Park, Hae Won  and
      Breazeal, Cynthia  and
      Sap, Maarten",
    editor = "Al-Onaizan, Yaser  and
      Bansal, Mohit  and
      Chen, Yun-Nung",
    booktitle = "Proceedings of the 2024 Conference on Empirical Methods in Natural Language Processing",
    month = nov,
    year = "2024",
    address = "Miami, Florida, USA",
    publisher = "Association for Computational Linguistics",
    url = "https://aclanthology.org/2024.emnlp-main.59/",
    doi = "10.18653/v1/2024.emnlp-main.59",
    pages = "1026--1046"
}

@inproceedings{hu-etal-2024-gentranslate,
    title = "{G}en{T}ranslate: Large Language Models are Generative Multilingual Speech and Machine Translators",
    author = "Hu, Yuchen  and
      Chen, Chen  and
      Yang, Chao-Han  and
      Li, Ruizhe  and
      Zhang, Dong  and
      Chen, Zhehuai  and
      Chng, EngSiong",
    editor = "Ku, Lun-Wei  and
      Martins, Andre  and
      Srikumar, Vivek",
    booktitle = "Proceedings of the 62nd Annual Meeting of the Association for Computational Linguistics (Volume 1: Long Papers)",
    month = aug,
    year = "2024",
    address = "Bangkok, Thailand",
    publisher = "Association for Computational Linguistics",
    url = "https://aclanthology.org/2024.acl-long.5/",
    doi = "10.18653/v1/2024.acl-long.5",
    pages = "74--90"
}

@article{Hu2024GRAGGR,
  title={GRAG: Graph Retrieval-Augmented Generation},
  author={Yuntong Hu and Zhihan Lei and Zhengwu Zhang and Bo Pan and Chen Ling and Liang Zhao},
  journal={ArXiv},
  year={2024},
  volume={abs/2405.16506},
  url={https://api.semanticscholar.org/CorpusID:270062608}
}

@inproceedings{wang-etal-2024-learning-plan,
    title = "Learning to Plan for Retrieval-Augmented Large Language Models from Knowledge Graphs",
    author = "Wang, Junjie  and
      Chen, Mingyang  and
      Hu, Binbin  and
      Yang, Dan  and
      Liu, Ziqi  and
      Shen, Yue  and
      Wei, Peng  and
      Zhang, Zhiqiang  and
      Gu, Jinjie  and
      Zhou, Jun  and
      Pan, Jeff Z.  and
      Zhang, Wen  and
      Chen, Huajun",
    editor = "Al-Onaizan, Yaser  and
      Bansal, Mohit  and
      Chen, Yun-Nung",
    booktitle = "Findings of the Association for Computational Linguistics: EMNLP 2024",
    month = nov,
    year = "2024",
    address = "Miami, Florida, USA",
    publisher = "Association for Computational Linguistics",
    url = "https://aclanthology.org/2024.findings-emnlp.459/",
    doi = "10.18653/v1/2024.findings-emnlp.459
        
        
        
        ",
    pages = "7813--7835",
}

@inproceedings{Cao2023InstructionMI,
  title={Instruction Mining: Instruction Data Selection for Tuning Large Language Models},
  author={Yihan Cao and Yanbin Kang and Chi Wang and Lichao Sun},
  year={2023},
  url={https://api.semanticscholar.org/CorpusID:264590782}
}

@article{Jiang2024KGFITKG,
  title={KG-FIT: Knowledge Graph Fine-Tuning Upon Open-World Knowledge},
  author={Pengcheng Jiang and Lang Cao and Cao Xiao and Parminder Bhatia and Jimeng Sun and Jiawei Han},
  journal={ArXiv},
  year={2024},
  volume={abs/2405.16412},
  url={https://api.semanticscholar.org/CorpusID:270062837}
 }

@article{Sun2023ThinkonGraphDA,
  title={Think-on-Graph: Deep and Responsible Reasoning of Large Language Model with Knowledge Graph},
  author={Jiashuo Sun and Chengjin Xu and Lumingyuan Tang and Sai Wang and Chen Lin and Yeyun Gong and Heung-yeung Shum and Jian Guo},
  journal={ArXiv},
  year={2023},
  volume={abs/2307.07697},
  url={https://api.semanticscholar.org/CorpusID:259936842}
}

@inproceedings{Zhang2019DIALOGPTL,
  title={DIALOGPT : Large-Scale Generative Pre-training for Conversational Response Generation},
  author={Yizhe Zhang and Siqi Sun and Michel Galley and Yen-Chun Chen and Chris Brockett and Xiang Gao and Jianfeng Gao and Jingjing Liu and William B. Dolan},
  booktitle={Annual Meeting of the Association for Computational Linguistics},
  year={2019},
  url={https://api.semanticscholar.org/CorpusID:207869708}
}

@article{Basyal2023TextSU,
  title={Text Summarization Using Large Language Models: A Comparative Study of MPT-7b-instruct, Falcon-7b-instruct, and OpenAI Chat-GPT Models},
  author={Lochan Basyal and Mihir Sanghvi},
  journal={ArXiv},
  year={2023},
  volume={abs/2310.10449},
  url={https://api.semanticscholar.org/CorpusID:264146201}
}

@article{Wu2024SparkRAAR,
  title={SparkRA: A Retrieval-Augmented Knowledge Service System Based on Spark Large Language Model},
  author={Dayong Wu and Jiaqi Li and Baoxin Wang and Honghong Zhao and Siyuan Xue and Yanjie Yang and Zhijun Chang and Rui Zhang and Li Qian and Bo Wang and Shijin Wang and Zhixiong Zhang and Guoping Hu},
  journal={ArXiv},
  year={2024},
  volume={abs/2408.06574},
  url={https://api.semanticscholar.org/CorpusID:271859627}
}

@article{Wu2024MedicalGR,
  title={Medical Graph RAG: Towards Safe Medical Large Language Model via Graph Retrieval-Augmented Generation},
  author={Junde Wu and Jiayuan Zhu and Yunli Qi},
  journal={ArXiv},
  year={2024},
  volume={abs/2408.04187},
  url={https://api.semanticscholar.org/CorpusID:271768853}
}

@article{li2025perception,
  title={Perception, Reason, Think, and Plan: A Survey on Large Multimodal Reasoning Models},
  author={Li, Yunxin and Liu, Zhenyu and Li, Zitao and Zhang, Xuanyu and Xu, Zhenran and Chen, Xinyu and Shi, Haoyuan and Jiang, Shenyuan and Wang, Xintong and Wang, Jifang and Huang, Shouzheng and Zhao, Xinping and Jiang, Borui and Hong, Lanqing and Wang, Longyue and Tian, Zhuotao and Huai, Baoxing and Luo, Wenhan and Luo, Weihua and Zhang, Zheng and Hu, Baotian and Zhang, Min},
  journal={arXiv preprint arXiv:2505.04921},
  year={2025}
}

\newpage
\appendix

\section{EXPERIMENT DETAILS}
\label{sec:EXPERIMENT DETAILS}
\subsection{Datasets}
To evaluate the performance of ORT on knowledge graph question and answer tasks, we conducted experiments on two multi-hop datasets (CWQ \cite{Talmor2018TheWA} and WebQSP \cite{Devlin2019BERTPO}). The questions in both datasets cover various domains, including people, places, events, etc. Due to the complexity of the questions, traditional question answering systems and search-based engines often struggle to provide valuable knowledge. FreeBase \cite{Bollacker2008FreebaseAC} serves as the background knowledge graph for both datasets, containing approximately 88 million entities, 20,000 relationships, and 126 million triples.

Similar to the datasets used in ROG, we extracted 3,531 question-answer pairs from the CWQ dataset as the test set, which includes 2,294,264 triples and 4,726 relationships. We also extracted 1,628 question-answer pairs from the WebQSP dataset as the test set, which includes 2,277,228 triples and 5,051 relationships. For details, see Table~\ref{tabel:introduce}.
\begin{table*}[htbp]
  \centering
  \normalsize
  \caption{The statistics of the used datasets.}
  \label{tabel:introduce}
  \renewcommand{\arraystretch}{1.2} % 调整行高
  \setlength{\tabcolsep}{12pt} % 调整列间距，这里稍微增大
  \begin{tabular}{c c c}  % 第一个列左对齐，后两列居中对齐
    \toprule
    \textbf{Datasets} & \textbf{Complex WebQuestions} & \textbf{WebQuestionSP} \\
    \midrule
    Domain & English General Q\&A & English General Q\&A\\  % 合并并居中
    KG dataset & FreeBase & FreeBase\\  % 合并并居中
    Question & 3531 & 1628 \\
    Node & 684846 & 781490 \\
    Triple & 2294264 & 2277228 \\
    Relationship & 4726 & 5051 \\
    \bottomrule
  \end{tabular}
\end{table*}

\subsection{Metrics}
\textbf{Accuracy} is the ratio of the number of correct predictions to the total number of predictions. The formula is as follows:
\begin{equation}
\text{Accuracy} = \frac{\sum_{i=1}^{N} \mathbb{I}(\hat{y}_i \in A_{\text{gold},i})}{N}
\end{equation}

\textbf{Hit@1} is whether the most probable prediction among the model's multiple outputs contains the ground truth. If yes, the Hit@1 score is 1; otherwise, the score is 0. Because our method has only one output, there is no need to select the prediction with the highest probability. For example, consider the question “What religion does India follow?” The correct answer is "Hinduism," and the model’s predicted answers are "Christianity, Hinduism, Islam." In this case, since "Hinduism" appears in the model's predicted answers and it is the correct answer, the Hit@1 score is 1. The formula is as follows:
\begin{equation}
Hit@1 = \mathbb{I}(\exists \hat{y}_i \in A_{\text{gold}})
\end{equation}

\textbf{Precision} is the ratio of the number of correct predictions to the total number of predictions. The formula is as follows:
\begin{equation}
Precision = \frac{\sum_{i=1}^{N} \mathbb{I}(\hat{y}_i \in A_{\text{gold},i})}{N_{\text{pred}}}
\end{equation}

\textbf{Recall} measures how many of the standard answers the model can correctly predict. The calculation method is the same as that of accuracy.

\textbf{F1 score} is the harmonic mean of precision and recall. The formula is as follows:
\begin{equation}
F1 = \frac{2 \times Precision \times Recall}{Precision + Recall}
\end{equation}

\subsection{Baselines}
Baselines are grouped into 12 baseline methods into 5 categories: 1) \textbf{Embedding-based methods}, 2) \textbf{Retrieval-augmented methods}, 3) \textbf{LLM}, 4) \textbf{LLM+KGs (Fine-tuned)}, and 5) \textbf{LLM+KGs (non-Fine-tuned)}. The detailed information for each baseline is as follows:

\textbf{Embedding-based methods}  
\begin{itemize}
    \item KV-MEM \cite{Miller2016KeyValueMN} employs a key-value memory network to store triples and performs multi-hop reasoning by iterating operations over memory.  
    \item NSM\cite{He2021ImprovingMK} uses a sequential model to mimic the multi-hop reasoning process.
\end{itemize}

\textbf{Retrieval-augmented methods}  
\begin{itemize}
    \item GraftNet \cite{Sun2018OpenDQ}  retrieves relevant subgraphs from knowledge graphs with entity linking.  
    \item SR+NSM \cite{Zhang2022SubgraphRE} introduces a relation-path retrieval mechanism to fetch subgraphs for multi-hop reasoning.  
    \item SR+NSM+E2E \cite{Zhang2022SubgraphRE} further adopts an end-to-end training strategy to jointly train the retrieval and reasoning modules of SR+NSM.
\end{itemize}

\textbf{LLM methods}  
\begin{itemize}
    \item GPT-4 is a large language model developed by OpenAI, renowned for its excellent performance across a wide range of natural language processing tasks.  
    \item DeepSeek-v3 \cite{DeepSeekAI2024DeepSeekV3TR} is an advanced model designed for deep reasoning and retrieval-augmented tasks, focusing on domain-specific knowledge extraction.  
    \item Qwen-max is a large model optimized for multilingual and multi-task learning, known for its strong capabilities in both generative and analytical tasks.
\end{itemize}

\textbf{LLM+KGs (Fine-tuned) methods}  
\begin{itemize}
    \item KD-COT \cite{Wang2023KnowledgeDrivenCE} retrieves relevant knowledge from KGs to formulate faithful reasoning plans for LLMs.  
    \item ROG \cite{Luo2023ReasoningOG} combines knowledge graphs (KGs) and large language models (LLMs) to achieve reliable and interpretable reasoning through a planning-retrieval-reasoning framework.
\end{itemize}

\textbf{LLM+KGs (non-Fine-tuned) methods}  
\begin{itemize}
    \item KG Retriever aims to find the shortest path between each pair of question entities, and then retrieves the final prompt from the KG to guide the LLM in answering the question. The key difference between MindMap and KG Retriever is that they do not use diverse multiple pieces of evidence in the LLM, nor do they ORT the evidence sources.  
    \item MindMap \cite{wen-etal-2024-mindmap} integrates knowledge graphs (KGs) and large language models (LLMs) by using KGs to provide explicit knowledge and reasoning paths, enhancing the LLM's reasoning ability and transparency while revealing the thought process of the LLM.
\end{itemize}

\section{Implementation Details}
\label{sec:Implementation Details}
\subsection{Label List Construction}
We utilize the datasets provided by RoG to conduct our experiments. The relations in the triples of the knowledge graph contain label information. For example, in the triple [Jamaica, meteorology.cyclone\_affected\_area.cyclones, Tropical Storm Keith], the entity ``Jamaica'' is assigned the label \textit{cyclone\_affected\_area}, while ``Tropical Storm Keith'' is labeled as \textit{cyclones}.

\subsection{Ontology Construction}
We traverse the dataset triples and extract entity label information in a normalized format, constructing abstract triples of the form [label, relation, label]. These abstract triples collectively form the knowledge graph ontology.

\subsection{Hyperparameter Settings}

The hyperparameters used in our experiments are as follows:

\begin{itemize}
  \item \textbf{Maximum reasoning hops:} \texttt{MAX\_POP = 5}. The maximum hop count for the WebQSP and CWQ datasets is 4, so the maximum reasoning hops are set to 5 to ensure sufficient reasoning depth.

  \item \textbf{Maximum number of neighbors:} \texttt{TOP\_K = 10}. Here, neighbors refer to entities in the \textit{Guided Answer Mining} phase that satisfy the \texttt{next\_label} constraint.
\end{itemize}

\section{ADDITIONAL RESULTS}
\label{sec:ADDTIONAL RESULTS}
We have added Accuracy, Precision, and Recall to further observe the Model Improvement Comparison between WebQuestionSP and ComplexWebQuestions. You can find the details in Table~\ref{tabel:detail}.

\section{PROMPTS}
\label{sec:PROMPTS}
\begin{figure}[t]
  \centering
  \includegraphics[width=\columnwidth]{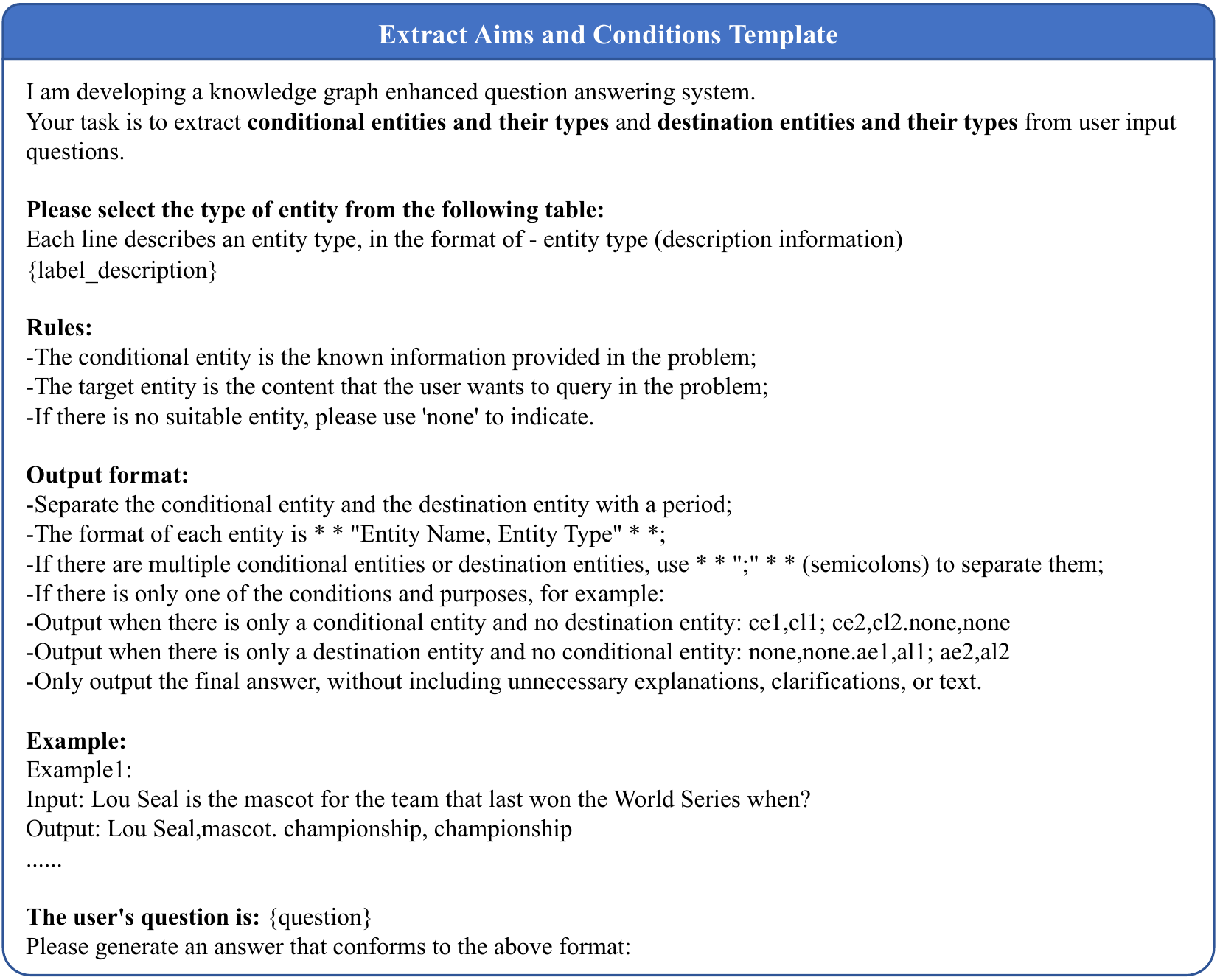}  % 设置图片宽度为两栏宽
  \caption{The prompt template for "Aim and Condition Recognition"}  % 图片标题
  \label{fig:extract aims and conditions template}  % 图片引用标签
\end{figure}

\begin{figure}[t]
  \centering
  \includegraphics[width=\columnwidth]{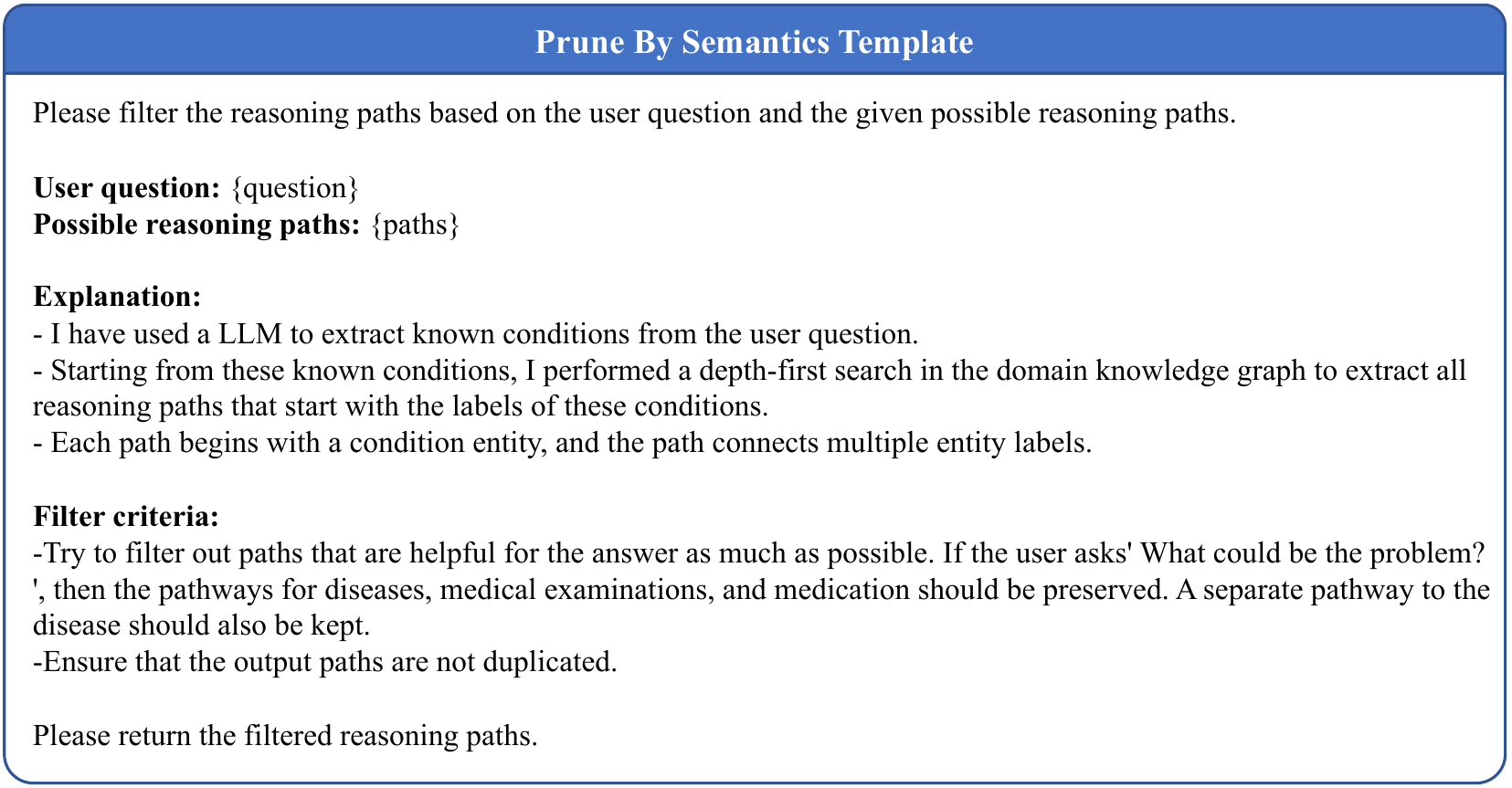}  % 设置图片宽度为两栏宽
  \caption{The prompt template for "Prune by Semantics"}  % 图片标题
  \label{fig:prune by semantics template}  % 图片引用标签
\end{figure}

\begin{figure}[t]
  \centering
  \includegraphics[width=\columnwidth]{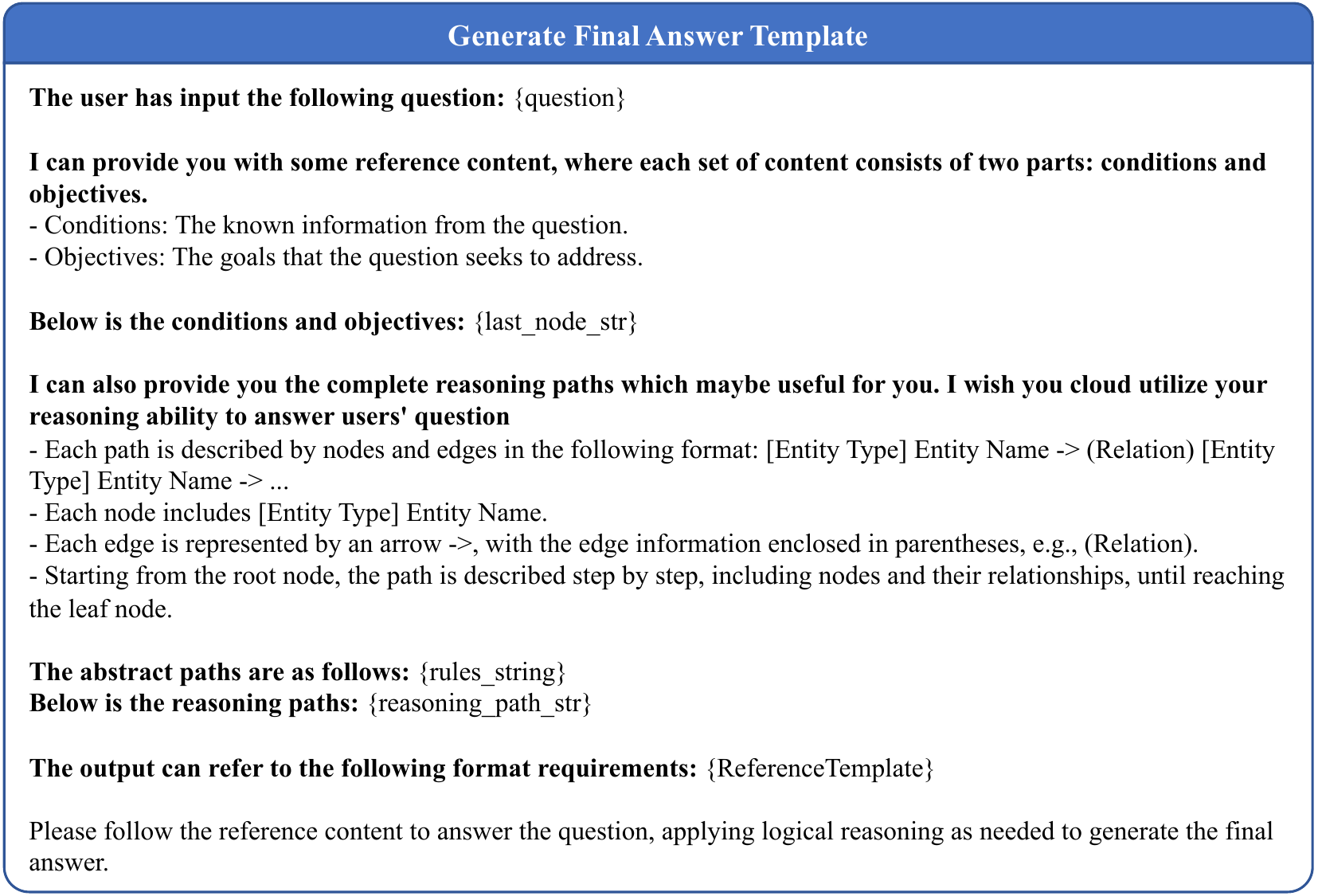}  % 设置图片宽度为两栏宽
  \caption{The prompt template for "Guided Answer Mining"}  % 图片标题
  \label{fig:generate answer template}  % 图片引用标签
\end{figure}

We demonstrate all the prompt templates used, including ``Aims and Conditions Recognition'', ``Prune by Semantics'' and ``Generate Final Answer With LLMs'', as shown in Figure~\ref{fig:extract aims and conditions template}, Figure~\ref{fig:prune by semantics template}, and Figure~\ref{fig:generate answer template}.

\section{CASES}
\label{sec:CASES}
We will present two cases in Table~\ref{table:case1} and Table~\ref{table:case2} to illustrate the process of our method.

\onecolumn

\begin{table*}[h]
\centering
\scriptsize
\caption{Detailed Experiment Results of Model Improvement Comparison between WebQSP and CWQ}
\label{tabel:detail}
\renewcommand{\arraystretch}{1.5} % 调整行间距
\begin{tabular}{l c c c c c | c c c c c}
\toprule
 \multirow{2}{*}{Method} & \multicolumn{5}{c|}{WebQSP} & \multicolumn{5}{c}{CWQ} \\
\cline{2-11}
 & Accuracy & Hit@1 & Precision & Recall & F1 & Accuracy & Hit@1 & Precision & Recall & F1 \\
\hline
GPT-4o & 43.29 & 61.79 & 61.07 & 43.29 & 43.56 & 33.61 & 38.20 & 36.56 & 33.61 & 32.87 \\
GPT-4o + ORT & 71.86 & 87.67 & 88.00 & 71.86 & 71.79 & 58.99 & 65.43 & 63.21 & 58.99 & 58.69 \\
GPT-4o + MindMap & 47.81 & 61.17 & 50.33 & 47.81 & 46.09 & 45.52 & 51.33 & 49.50 & 45.52 & 44.84 \\
GPT-4o + KG Retriever & 44.72 & 60.15 & 46.64 & 44.72 & 42.44 & 40.82 & 46.67 & 45.77 & 40.82 & 41.14 \\
\hline
QWen-max & 41.39 & 59.00 & 55.65 & 41.39 & 40.04 & 31.66 & 36.42 & 32.29 & 31.66 & 29.45 \\
QWen-max + ORT & 74.30 & 88.14 & 81.00 & 74.30 & 71.73 & 61.61 & 67.87 & 58.97 & 61.61 & 57.75 \\
QWen-max + MindMap & 46.09 & 59.46 & 46.57 & 46.11 & 43.31 & 40.59 & 45.50 & 44.04 & 40.59 & 40.35 \\
QWen-max + KG Retriever & 42.23 & 57.16 & 43.93 & 42.25 & 39.91 & 40.03 & 45.00 & 42.31 & 40.03 & 38.99 \\
\hline
DeepSeek-v3 & 46.55 & 64.00 & 56.91 & 46.55 & 43.87 & 36.40 & 41.12 & 36.42 & 36.40 & 33.80 \\
DeepSeek-v3 + ORT & 74.51 & 89.43 & 80.92 & 74.51 & 71.83 & 66.03 & 72.91 & 65.57 & 66.03 & 62.63 \\
DeepSeek-v3 + MindMap & 50.68 & 64.92 & 50.10 & 50.68 & 47.14 & 44.07 & 48.83 & 46.79 & 44.07 & 43.30 \\
DeepSeek-v3 + KG Retriever & 47.88 & 63.01 & 46.18 & 47.88 & 42.87 & 42.95 & 47.67 & 41.87 & 42.95 & 40.20 \\
\bottomrule
\end{tabular}
\end{table*}

\newpage
\begin{table}[htbp]
    \centering
    \caption{Two cases for better understanding of our method: Case 1.}
    \label{table:case1}
    \begin{tabular}{p{15cm}}
        \toprule
        % \textbf{Case1} \\
        % \midrule
        \textbf{Question:} \\
        Lou Seal is the mascot for the team that last won the World Series when? \\
        \vspace{2mm} % 插入空行
        \textbf{Aims:} \\
        \texttt{[["championship", "championship"]]} \\
        \vspace{2mm} % 插入空行
        \textbf{Conditions:} \\
        \texttt{[["Lou Seal", "mascot"]]} \\
        \vspace{2mm} % 插入空行
        \textbf{Rule\_Paths:} \\
        \texttt{mascot -> game -> season -> championship} \\
        \texttt{mascot -> team -> season -> championship} \\
        \texttt{mascot -> team -> championship} \\
        \texttt{mascot -> school -> team -> championship} \\
        \texttt{mascot -> brand -> team -> championship} \\
        \texttt{mascot -> team -> league -> championship} \\
        \texttt{mascot -> team -> relationship -> championship} \\
        \texttt{mascot -> brand -> relationship -> championship} \\
        \texttt{mascot -> game -> event -> championship} \\
        \texttt{mascot -> team -> event -> championship} \\
        \vspace{2mm} % 插入空行
        \textbf{Selected\_Rule\_Paths:} \\
        \texttt{mascot -> team -> championship} \\
        \texttt{mascot -> team -> event -> championship} \\
        \texttt{mascot -> team -> season -> championship} \\
        \vspace{2mm} % 插入空行
        \textbf{Reasoning\_Paths:} \\
        reasoning path 1: \texttt{[mascot] Lou Seal -> team [team] San Francisco Giants -> champion [championship] 2010 World Series} \\
        % \vspace{1mm} % 插入空行
        reasoning path 2: \texttt{[mascot] Lou Seal -> team [team] San Francisco Giants -> championship [championship] 2014 World Series} \\
        % \vspace{1mm} % 插入空行
        reasoning path 3: \texttt{[mascot] Lou Seal -> team [team] San Francisco Giants -> champion [championship] 2012 World Series} \\
        \dots \\
        
        reasoning path 82: \texttt{[mascot] Lou Seal -> team\_mascot [team] San Francisco Giants -> league [season] m.0crt4b6} \\
        % \vspace{1mm} % 插入空行
        reasoning path 83: \texttt{[mascot] Lou Seal -> team\_mascot [team] San Francisco Giants -> team [season] National League West -> championship [championship] National League Division Series} \\
        \vspace{2mm} % 插入空行
        
        \textbf{Final\_Answer:} \\
        2014 World Series \\
        % \vspace{2mm} % 插入空行
        \bottomrule
    \end{tabular}
\end{table}

\newpage
\begin{table}[htbp]
    \centering
    \caption{Two cases for better understanding of our method: Case 2.}
    \label{table:case2}
    \begin{tabular}{p{15cm}}
        \toprule
        % \textbf{Case2} \\
        % \midrule
        
        \textbf{Question:} \\
        What is the predominant religion where the leader is Ovadia Yosef? \\
        \vspace{2mm} % 插入空行
        \textbf{Aims:} \\
        \texttt{[["religion", "religion"]]} \\
        \vspace{2mm} % 插入空行
        \textbf{Conditions:} \\
        \texttt{[["Ovadia Yosef", "person"]]} \\
        \vspace{2mm} % 插入空行

        \textbf{Rule\_Paths:} \\
        \texttt{person -> religion} \\
        \texttt{person -> party -> celebrity -> religion} \\
        \texttt{person -> language -> region -> religion} \\
        \texttt{person -> title -> membership -> religion} \\
        \texttt{person -> group -> membership -> religion} \\
        \texttt{person -> leadership -> organization -> religion} \\
        \texttt{person -> child -> organization -> religion} \\
        \texttt{person -> location -> organization -> religion} \\
        \texttt{person -> parent -> organization -> religion} \\
        \texttt{person -> title -> leader -> religion} \\
        \texttt{person -> leadership -> leader -> religion} \\
        \texttt{person -> location -> choice -> religion} \\
        \vspace{2mm} % 插入空行
        \textbf{Selected\_Rule\_Paths:} \\
        \texttt{person -> leadership -> leader -> religion} \\
        \texttt{person -> title -> leader -> religion} \\
        \texttt{person -> leadership -> organization -> religion} \\
        \vspace{2mm} % 插入空行

        \textbf{Reasoning\_Paths:} \\
        reasoning path 1: \texttt{[person] Ovadia Yosef ->  leader [leadership] m.048bcbz ->  leader [leader] Ovadia Yosef ->  religion [religion] Judaism} \\
        % \vspace{2mm} % 插入空行
        reasoning path 2: \texttt{[person] Ovadia Yosef ->  leader [leadership] m.048bcbz ->  leader [leader] Ovadia Yosef ->  religion [religion] Haredi Judaism} \\
        % \vspace{2mm} % 插入空行
        reasoning path 3: \texttt{[person] Ovadia Yosef ->  leader [leadership] m.048bcbz ->  religious\_leadership [leader] Ovadia Yosef ->  religion [religion] Judaism} \\
        % \vspace{2mm} % 插入空行
        reasoning path 4: \texttt{[person] Ovadia Yosef ->  leader [leadership] m.048bcbz ->  religious\_leadership [leader] Ovadia Yosef ->  religion [religion] Haredi Judaism} \\
        \dots \\
        reasoning path 20: \texttt{[person] Ovadia Yosef ->  religious\_leadership [leadership] m.048bcbz ->  religious\_leadership [organization] Ovadia Yosef ->  religion [religion] Haredi Judaism} \\
        % \vspace{2mm} % 插入空行
        reasoning path 21: \texttt{[person] Ovadia Yosef ->  religious\_leadership [leadership] m.048bcbz ->  organization [organization] Chief Rabbinate of Israel} \\
        \vspace{2mm} % 插入空行

        \textbf{Final\_Answer:} \\
        Judaism \\
        % \vspace{2mm} % 插入空行
        \bottomrule
    \end{tabular}
\end{table}

\end{document}